\documentclass[lettersize,journal]{IEEEtran}
\usepackage{amsmath,amsfonts}
\usepackage{algorithmic}
\usepackage{algorithm}
\usepackage{array}
\usepackage[caption=false,font=normalsize,labelfont=sf,textfont=sf]{subfig}
\usepackage{textcomp}
\usepackage{stfloats}
\usepackage{url}
\usepackage{verbatim}
\usepackage{graphicx}
\usepackage{cite}
\begin{document}

\title{Improving Location-based Thermal Emission Side-Channel Analysis Using Iterative Transfer Learning}

\author{Tun-Chieh Lou, Chung-Che Wang, Jyh-Shing Roger Jang, Henian Li, Lang Lin, and Norman Chang}



\maketitle

\begin{abstract}
This paper proposes the use of iterative transfer learning applied to deep learning models for side-channel attacks. Currently, most of the side-channel attack methods train a model for each individual byte, without considering the correlation between bytes. However, since the models’ parameters for attacking different bytes may be similar, we can leverage transfer learning, meaning that we first train the model for one of the key bytes, then use the trained model as a pretrained model for the remaining bytes. This technique can be applied iteratively, a process known as iterative transfer learning. Experimental results show that when using thermal or power consumption map images as input, and multilayer perceptron or convolutional neural network as the model, our method improves average performance, especially when the amount of data is insufficient.
\end{abstract}


\section{Introduction}

\IEEEPARstart{W}{ith} advancements in technology, the manufacturing processes of chips in electronic devices have continuously improved, but security concerns have also become increasingly critical. Among potential threats, the side-channel attack (SCA) is particularly notable due to its ability to efficiently and effectively extract encrypted keys and other hidden information by leveraging physical data leaks, such as power consumption or temperature variations, during encryption operations.

In recent years, machine learning has been widely applied to SCAs~\cite{hettwer2020applications}, where the attack can be framed as a classification problem. Models use physical data, such as power consumption or temperature changes, to predict encryption keys. Taking the example of breaking AES-128~\cite{AES} (Advanced Encryption Standard-128) in this paper, the observed physical data are fed into the model, which predicts the value of each byte after its transformation through the substitution box (S-box).

Among these machine learning methods, MultiLayer Perceptron (MLP) and Convolutional Neural Network (CNN) have been the focus of related studies. Yang et al.~\cite{yang2012back} conducted one of the earliest studies using MLP for side-channel attacks, employing regression analysis to predict the S-box byte outputs. Martinasek and Zeman~\cite{martinasek2013innovative} approached the problem differently by categorizing the S-box byte outputs into 256 labels and reframing the side-channel attack as a classification task. Maghrebi, Portigliatti and Prouff~\cite{maghrebi2016breaking} utilized CNNs to break AES encryption, demonstrating that CNNs achieve higher success rates than MLP and other models. Benadjila et al.~\cite{benadjila2020deep} further tested the success rates of CNNs and various other models, while also introducing the ASCAD database, providing a public benchmark for related research. In our study, we also evaluate the success rates of different models and propose a public dataset for comparison.

However, training neural networks typically requires a large amount of data. To improve data efficiency, transfer learning has been introduced to solve the problem of side-channel attacks. Thapar, Alam, and Mukhopadhyay~\cite{thapar2021deep}, along with Yu et al.~\cite{yu2021cross}, proposed pretraining using data from different devices. Garg and Karimian~\cite{garg2021leveraging} attempted to use pretrained models from general image recognition tasks, such as InceptionV3 or VGG16, as a foundation. However, other studies~\cite{hettwer2020encoding} have pointed out that the effectiveness of pretrained models can be inconsistent. In this study, since the data originates from a single device and the image characteristics differ significantly from those used in general image recognition models, we consider the correlations between different bytes and apply transfer learning to transfer knowledge across the 16 bytes in the model's output.


The rest of this paper is organized as follows. Section~\ref{sec:data} introduces our dataset, Section~\ref{sec:method} describes our methods for SCA, Section~\ref{sec:exp} shows the experimental results, and Section~\ref{sec:con} concludes this paper and addresses possible future work.

\section{Dataset}
\label{sec:data}
Our dataset is build based on the calculation process of AES encryption~\cite{AES}. The algorithm of AES encryption~\cite{AES} is composed of an initial round of \textit{AddRoundKey} operation, 9 (or 11, 13) rounds of \textit{SubBytes}, \textit{ShiftRows}, \textit{MixColumns}, and \textit{AddRoundKey} operation, and a final round of \textit{SubBytes}, \textit{ShiftRows}, and \textit{AddRoundKey} operation. For each 128-bit plain text, the simulated~\cite{Wen2022} power consumption or thermal map of the first \textit{AddRoundKey} and \textit{SubBytes} operation of a certain chip, and the output of the first \textit{SubBytes} operation is generated. The software for simulation is RedHawk-SC, and the key used for encryption is fixed as $\{0x00, 0x11, 0x22, ..., 0xFF\}$. The power consumption or thermal map is used as our model input, and the output of the first \textit{SubBytes} operation is used as the model output. The dataset is composed of 20,000 input-output pairs, together with the 16 expert-specified points of interest (POIs) respectively for the 16 output bytes.

\begin{figure}[!t]
\centering
\includegraphics[width=2.5in]{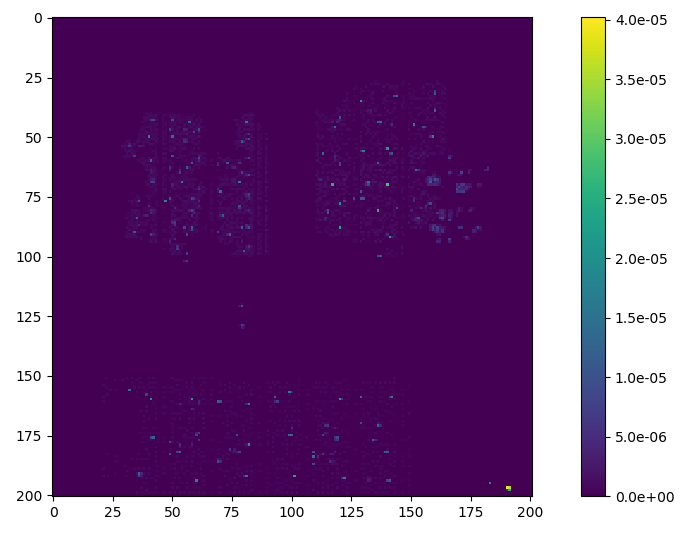}
\caption{An example of the power consumption map.}
\label{fig_power}
\end{figure}

\begin{figure}[!t]
\centering
\includegraphics[width=2.5in]{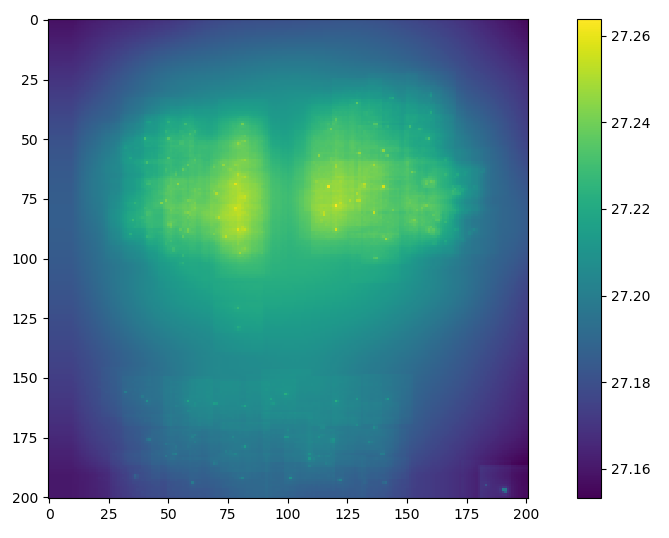}
\caption{An example of the thermal map.}
\label{fig_thermal}
\end{figure}

Fig.~\ref{fig_power} is an example of the power consumption map. The height and width of the image is 201, where each pixel represents the power consumption (in watts) of a 100 square micrometer area during the first \textit{AddRoundKey} and \textit{SubBytes} operation of the AES encryption. Since most of the pixels are with zero value, the power consumption map can be used for selecting important regions. Fig.~\ref{fig_thermal} is an example of the thermal map. The height and width of the image are also 201, where each pixel represents the average temperature (in °C) of a 100 square micrometer area during AES encryption. Since heat spreads outward through the medium, each pixel's temperature is affected by its nearby pixels. Therefore, the pattern of the thermal map looks different from the power consumption map.

\section{Methods}
\label{sec:method}
\subsection{Data Preprocessing and Progressive Feature Selection}

\begin{figure}[!t]
\centering
\includegraphics[width=2.5in]{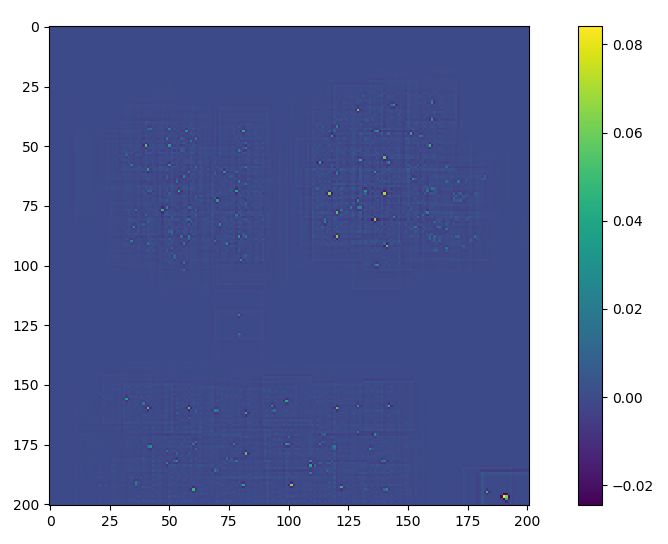}
\caption{An example of the filtered thermal map.}
\label{fig_thermal_filter}
\end{figure}

Since using all pixels as input features, which has 40,401 dimensions in our setting, may not be suitable for some classifiers including Multilayer Perceptron (MLP) and Support Vector Machine (SVM), data preprocessing and feature selection stages need to be invoked. A Laplacian filter is first applied to the thermal map to cancel the coupling effect. Fig.~\ref{fig_thermal_filter} is an example of the filtered thermal map, where the original input is Fig.~\ref{fig_thermal}. We did try other filters including Sobel filter and Prewitt filter, but Laplacian filter leads to the best results in the preliminary experiments.

Due to the intuition that a position with higher variation may be more important, we then calculate the standard deviation for each pixel for the filtered thermal map in the training data, and the pixels with top-200 highest standard deviations are then used for one-pass ranking and sequential forward feature selection. In these feature selection stages, logistic regression is used for obtaining the inside prediction results, and the cross-entropy loss is used as the criterion for selecting helpful features. The one-pass ranking method select 50 features out of the top-200 pixels with highest standard deviations for each target byte, and the sequential forward method select 10 features out of the 50 features for each byte.

\begin{figure}[!t]
\centering
\includegraphics[width=2.5in]{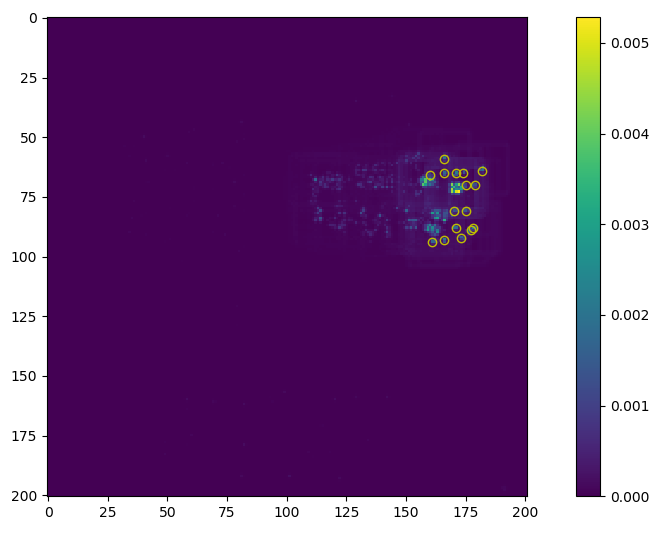}
\caption{An example of the calculated map of standard deviation using a random split training set, together with the 16 POIs.}
\label{fig_thermal_filter_std}
\end{figure}

Fig.~\ref{fig_thermal_filter_std} shows an example of the calculated map of standard deviation using a random split training set, together with the 16 expert-specified POIs. The pixels with top-200 highest standard deviation are gathered at the top-right corner, which are very close with the 16 POIs.

\subsection{Iterative Transfer Learning}

While most of researches trained one model for one byte of the encryption key, which did not consider the relevance between each byte, this research invokes transfer learning. Transfer learning is usually used as training on large-scale data in general domain, and fine-tune on a smaller set in a more specific domain. In this research, transfer learning is used iteratively, where the model trained for the first byte is used as the pretrained model for the second byte, and so on. This manner let the deep learning models crack the encryption successfully even when the dataset is not large. Our preliminary experiments shows that the model performance converges within two iterations, and the effectiveness of the order of bytes is slight.

\section{Experimental Results}
\label{sec:exp}
\subsection{Experimental Setup}

Since it is difficult to crack the AES-128 encryption using a single input, the model performance is evaluated as ``how many inputs should be used to crack the AES-128 encryption''. For the $b$-th byte of the key, the products of probabilities of the 256 possible values is calculated using the output of the first $i$ samples in the test set, and the $rank_{b,i}$ can be defined as the rank (0 to 255, where 0 means the highest rank, and 255 means the lowest rank) of the intended value out of the 256 possible values of the $b$-th byte when considering the first $i$ samples in the test set. The Measurement-to-disclosure (MTD) of the $b$-th byte is therefore defined as the minimal $i$ such that $rank_{b,j}$ is zero for all $i \ge j$, and the average and the worst MTDs over all the 16 bytes of the key are reported as the evaluation result, where a lower value indicates the better performance. Note that since the $rank_{b,i}$ and MTD are affected by the order of the test samples, the test data is repermutated 100 times for each byte to obtain an average result.

Experiments are conducted on a personal computer with Ubuntu 18.04, Intel i5-9400F CPU, and NVIDIA GeForce RTX 2070 SUPER GPU. Hyperparameters or structures of different model types are listed below:
\begin{itemize}
\item Random forest (RF): the scikit-learn\cite{scikit-learn} implementation is used. The entropy function is used as the quality measurement of a split, and other parameters are left as default.
\item Logistic regression (LR): the scikit-learn\cite{scikit-learn} implementation is used. The maximum number of iterations is set to 1,000,000, and other parameters are left as default.
\item SVM: the scikit-learn\cite{scikit-learn} implementation is used. Linear kernel is used and probability estimation is enabled. Other parameters are left as default.
\item MLP: three linear layers is used, each followed by batch normalization, the Mish activation function, and a dropout layer, except for the final linear layer. The first two linear layers have an output shape of 20, while the last one expands the output to 256.
\item CNN: three convolutional layers with batch normalization, Mish activation, and average pooling are first used. The number of channels reduces from 64 to 32 to 16 across the layers, with 3x3 kernels for the convolutional layers and 2x2 kernels for the pooling layers. The output is then flattened and passed through three linear layers, each with batch normalization, Mish activation, and dropout. The first two linear layers output 64 units each, while the final layer expands the output to 256 units.
\end{itemize}
For the deep learning models, the Ranger optimizer and cross-entropy loss are used. Early stopping is applied if the validation loss does not decrease for 100 epochs.

\subsection{Results: Comparison of Model Types}

We first respectively use 13,600, 3,400, and 3,000 thermal map images for training, validation, and test data for comparing the MTD for different models. Table~\ref{tab:cmp-models} shows the evaluation results, where the input dimensions are respectively 201x201 for CNN and CNN with ITL, 10 for MLP with ITL, and 200 for other methods, and the average rank is calculated as the average of $rank_{b,3000}$ for all the bytes. As shown in the table, RF is not able to crack AES-128, and most of the machine learning methods performs better then CPA. When using the proposed ITL on MLP and CNN, the MTD is further reduced, showing the effectiveness of our ITL.

\begin{table}[!t]
\caption{Comparing of Different Models\label{tab:cmp-models}}
\centering
\begin{tabular}{crrr}
\hline
Method~\textbackslash~Metric & Average MTD & Worst MTD & Average Rank \\
\hline
CPA & 1,402.0 & 2750 & 0 \\
RF & N/A & N/A & 46 \\
LR & 334.0 & 937 & 0 \\
SVM & 647.2 & 2990 & 0 \\
MLP & 54.7 & 111 & 0 \\
CNN & 125.0 & 233 & 0 \\
MLP with ITL & 53.8 & 88 & 0 \\
CNN with ITL & 82.7 & 164 & 0 \\
\hline
\end{tabular}
\end{table}

\subsection{Results: Comparison of Data Sizes}

\begin{figure}[!t]
\centering
\includegraphics[width=3in]{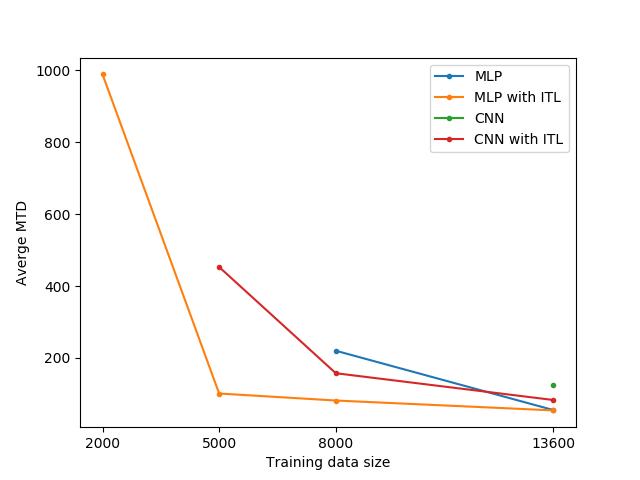}
\caption{MTDs of MLP and CNN with or without ITL when the training data size varies.}
\label{fig_less_data}
\end{figure}

We then investigate the performance of ITL when the training data size is reduced. The training data sizes are reduced to 2,000, 5,000, or 8,000 thermal map images, with the validation data size set to one-quarter of the corresponding training data size, and the test data size fixed at 3,000 images. Fig.~\ref{fig_less_data} shows the MTDs of MLP and CNN with or without ITL as the training data size varies. The absence of a point indicates that a particular model is unable to crack all the bytes of the encryption key with a certain size of training data. As the training data size decreases, the MTD increases, and MLP and CNN fail to crack all the bytes when the training data size is small. However, after applying ITL, the MTDs decrease for the same training data sizes, allowing the encryption to be cracked even with smaller training data sizes.

\subsection{Results: Performance Using Power Consumption Map}


The performance of using the power consumption map is also compared, with the MTDs shown in Table~\ref{tab:power}. Since the power consumption map is not affected by the coupling effect, important features are concentrated in a few pixels, enabling RF to crack the encryption. Most methods, including CPA and SVM, achieve lower MTDs than when using the thermal map. However, because CNN does not focus on specific pixels, its MTD using the power consumption map is higher than when using the thermal map. After applying ITL to CNN, the MTD is significantly reduced, once again demonstrating the effectiveness of our ITL.

\begin{table}[!t]
\caption{Average MTDs of Using Power Consumption Map\label{tab:power}}
\centering
\begin{tabular}{cr}
\hline
Method & MTD \\
\hline
CPA & 423.0 \\
RF & 1037.0 \\
LR & 313.0 \\
SVM & 111.0 \\
MLP & 54.4 \\
CNN & 204.0 \\
MLP with ITL & 52.0 \\
CNN with ITL & 56.0 \\
\hline
\end{tabular}
\end{table}

\section{Conclusions and Future Work}
\label{sec:con}
This paper proposes ITL for cracking the AES encryption, which iteratively trained models on each byte of the encryption key. Experimental results on thermal map show that applying ITL on MLP and CNN leads to better performance, even when the training data size is smaller. Besides, experiments using the power consumption map show improved results except CNN without ITL since the characteristic of the power consumption map is not beneficial for CNN, but applying ITL still lead to improved performance, showing the effectiveness of our proposed ITL for cracking the AES encryption.

One of the future research directions could be investigating the performance using different types of physical leakage information for the model input. We would also like to try using multitask learning to reduce the training time.



\bibliographystyle{IEEEbib}
\bibliography{refs}

\vfill

\end{document}